# Autonomous Navigation in Complex Environments

Andrew Gerstenslager, Jomol Lewis, Liam McKenna, and Poorva Patel

## I. ABSTRACT

**This paper explores the application of CNN-DNN network fusion to construct a robot navigation controller within a simulated environment. The simulated environment is constructed to model a subterranean rescue situation, such that an autonomous agent is tasked with finding a goal within an unknown cavernous system. Imitation learning is used to train the control algorithm to use LiDAR and camera data to navigate the space and find the goal. The trained model is then tested for robustness using Monte-Carlo.**

## II. INTRODUCTION

Complex environments such as subterranean caverns, mines, rubble, and other confined spaces are hazardous for work such as welding, search and rescue, etc. The risk to human life when operating under these circumstances can be immense due to the lack of visibility, breathable atmosphere, and communications [1]. It is in this area that we hope to apply autonomous decision making to reduce human risk, increase robotic capability, and maintain effectiveness of operations.

The aim of this work is to apply autonomous agents to tasks such as search and rescue, environmental inspection, counterterrorism, and mapping in complex environments. Robots have already proven to be effective in reducing the risk to human life by being used as a liaison between the human controller and the task, such as bomb disposal [2]. However, robotics teams have struggled to apply their systems effectively in complex systems.

Many challenges have been identified so far with the use of robotic systems in complex environments: the primary hardships include low visibility, lack of communication, minimal preliminary information, lack of GPS, and minimal human intervention [3]. These problems increase the difficulty humans face while attempting to maintain direct control of robotic agents during operations.

Humans are able to navigate through complex environments through the use of a variety of sensors and learned behaviors, allowing them to make complex, intelligent decisions, as a result. To address potential problems that a robot may face in a complex environment while emulating human decision making, we propose an intelligent control architecture to allow an autonomous agent to operate without significant human intervention. Using deep learning (DL) as a tool helped to generate a mapping from sensor data to control inputs, which allowed for autonomous mapping and navigation for an agent in a complex environment. Many similar platforms use a wide sensor suite to gather a large number of various data samples during operation, which we will leverage with the use of a multi-modal deep network to map sensor input to a control output.

## III. RELATED STUDIES

The survey by Murphy et al. [1] emphasized that mine rescue scenarios demand robots tailored with navigation, localization, and mapping capabilities designed for underground environments, particularly when human-robot interchangeability is not a feasible option.

Dang et al. [4] noted LIDAR's illumination-agnostic nature for effective underground localization and mapping. In their work, they introduce an autonomy framework for aerial robots in underground mine rescue missions. Their approach integrates multi-modal localization, a bifurcated exploration planner, and the YOLO network model for object detection. The objects' 3D locations are estimated through fused observations using binary Bayes filters. Field deployment in a real underground mine validates the method. In [5], they explored the challenge of enabling intelligent robots to navigate and avoid obstacles by leveraging multimodal fusion methods. They combine deep reinforcement learning (DRL) with various sensors, such as depth images, RGB images, and LiDAR, to enhance perceptual and decision-making capabilities. Their proposed model uses both, depth images and pseudo-LiDAR data from a single RGB-D camera, demonstrating the power of multimodal fusion for obstacle avoidance and navigation, especially in previously unseen scenarios.

In Ross et al. [6], an agent learns to play a racing game from visual data. A teacher plays the game with a controller, and the controller's input, along with the game's video stream, is used to create a training dataset. The video stream is stored as raw pixels and fed into a neural network with 1296 input nodes, a 32-node hidden layer, and 15 output nodes. In Chernova et al. [7], the study discusses a robot navigating a maze with sensory inputs, using an IR sensor and a teacher's controller. The IR sensor provides object proximity information. Rather than encoding expert knowledge procedurally, the robot maps sensor data to controller inputs directly. The study introduces a teaching-by-demonstration approach based on Gaussian mixture models, reducing the number of required demonstrations by allowing the agent to actively request and represent relevant training data.

The study carried out Hussein by et al. [8] employed a CNN model for object navigation, highlighting its effectiveness in autonomous navigation policy learning from demonstrations. Their Results showcased strong performance, with a 96.2% success rate in reaching the flag and moderate success rates of 53.1% for reaching objects and 40.70% for following lines. These outcomes demonstrate CNNs' ability to extract meaningful features from raw 3D environment images and learn policies from demonstrations. Additionally, the study underscored the value of active learning in enhancing learned policies with limited samples.

Cai et al [9] used a cross-modal fusion network to handle sensor noise and create a versatile navigation policy within a teacher-student distillation framework. The teacher develops an expert policy in an ideal environment, and the student emulates it while navigating with sensor noise. Their approach combines laser signal and map information to address pose estimation noise, improving generalization across various working conditions. Their model achieved an 80.2% success rate in a novel environment, defined as the average number of episodes where the robot successfully reached five goals in sequence within 60 seconds.

Colas et al. [10] proposed a path planning method to allow ground robots to navigate in 3D environments. This system is tailored for static environments, where the map is loaded prior to path planning and remains unchanged. Path planning is conducted using D*-Lite, enabling efficient replanning in reaction to cost adjustments or robot movement. Utilizing point cloud data, they assessed the system's performance in both basic 2D and challenging 3D staircase environments, showcasing the robot's ability to climb stairs.

Nguyen et al. [3] introduced NMFNet, an end-to-end, real-time deep learning framework designed for autonomous navigation within intricate environments, such as collapsed urban areas or natural caves. This model comprises three distinct branches dedicated to processing depth, RGB image, and point cloud data, allowing it to seamlessly translate sensory input data into precise steering commands. A comparative assessment against leading models in the field, including DroNet [11], Inception-V3 [12], and VariationNet [13], revealed the remarkable performance superiority of NMFNet. Notably, NMFNet achieved a substantially lower Root Mean Square Error (RMSE) score of 0.389, outperforming DroNet at 0.756 [11], Inception-V3 at 1.16 [12], and VariationNet at 1.436 [13].

The DARPA-organized Subterranean Challenge aimed to advance technologies for exploring challenging underground environments. Tranzatto et al. [14] discussed several robots that participated in the 2021 competition finals. Team Coordinated Robotics employed a combination of custom wheeled, tracked, and flying robots. They made use of various open-source mapping frameworks for ground vehicles and aerial robots adapting them to specific sensor configurations. To detect artifacts, they utilized the YOLOv4 network with onboard camera images, as well as gas sensors and Bluetooth detection.

Team MARBLE utilized Spot-legged robots and Husky wheeled robots during the Final Event. Their SLAM pipeline was based on the tightly-coupled LiDAR-inertial odometry framework as described in [15] For volumetric mapping, they employed a modified version of [Hornung et al., 2013], which facilitated efficient map sharing and incorporated semantic information related to traversability and stairs.

## IV. METHODOLOGY

### A. Model Architecture

In our proposed architecture, as depicted in Figure 1, we employ a multi-modal deep learning model that leverages LiDAR sensor data and image data. The architecture is designed to process and learn from both types of data concurrently, thereby enhancing its performance in environmental navigation tasks.

The architecture comprises two distinct pathways:

**LiDAR Sensor Data Pathway**: LiDAR sensor data is processed through a neural network architecture. The input layer accepts an input size of 20 datapoints, each representative of respective distances recorded by the LiDAR. This input is followed by two fully connected layers, each with 64 neurons. These layers allow the network to discern complex patterns from raw LiDAR datasets through feature extraction.

**Image Data Pathway**: The image data is processed through a Convolutional Neural Network (CNN). Initially, the data passes through two convolutional (Conv2D) layers with 8 and 16 filters, respectively, both with a kernel size of 4x4. These convolutional layers are followed by a flattening step to convert the 2D feature maps into a 1D feature vector.

**Concatenation**: The flattened image outputs of the image data pathway and the DNN processed outputs of the LiDAR sensor data pathway are concatenated to combine the features extracted from both data types. This concatenated vector is then passed through a series of dense layers with 128, 64, 32, and 16 neurons respectively. The final output layer uses a softmax activation function over the model's three output neurons to produce a probability distribution over the three possible actions. This serves as the control vector for autonomous navigation.

The model is compiled with the Stochastic Gradient Descent (SGD) optimizer. The use of categorical crossentropy loss function aims to improve accuracy during training.

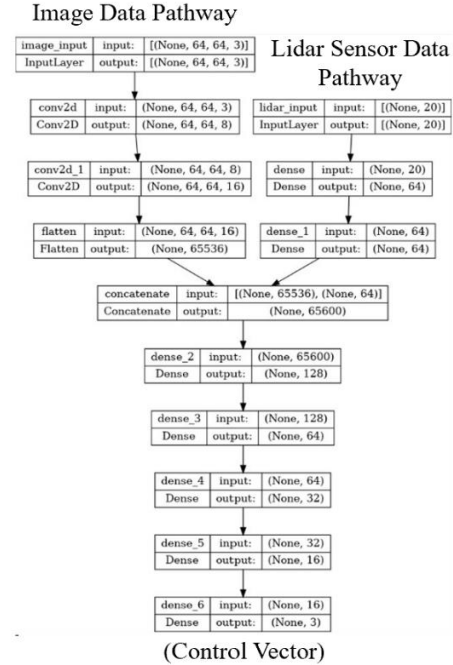

*Figure 1: Architecture of the multi-modal Deep Learning model utilizing Lidar Sensor and image data for autonomous Navigation.*

This architecture design ensures a comprehensive utilization of both LiDAR sensor and image data, potentially leading to improved model performance.

### B. Data Collection and Preprocessing

This autonomous deep learning project operated exclusively within a virtual environment and focused on the development of a sophisticated multimodal system for autonomous navigation. The system utilized image and LiDAR sensor data, both of which were generated through simulation.

The simulated camera contributed a layer of information to the multimodal system by capturing image data as 64x64x3-pixel images, simulating a robust 3D representation of the environment.

The strategically placed simulated LiDAR sensor atop the agent played an integral role in emulating the real-world functionalities of LiDAR technology by providing detailed depth information about the surroundings. This capability enabled the agent to perceive the environment in three dimensions by emitting laser beams and measuring their return time after reaching object surfaces.

The images, obtained alongside LiDAR data, offered visual insights that directly complemented the depth information provided by LiDAR.

The simulated environment was initially constructed in a 2D space, forming the foundation for a dynamic simulation. To enhance environmental representation, ray casting was used to generate 3D virtual camera images. This three-dimensional perspective, captured by the simulated camera, served as the input for the image data pathway in the multimodal system.

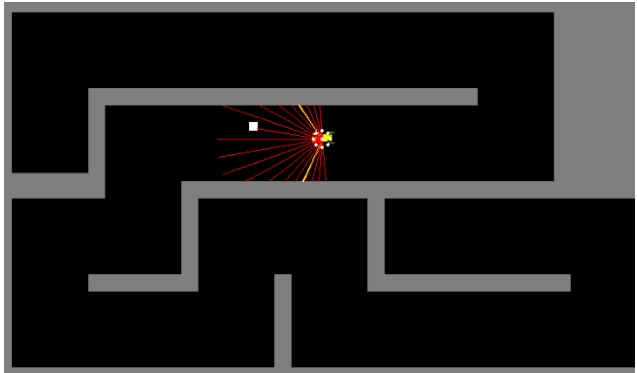

*Figure 2: The foundational 2D space representing the initial operating environment of the autonomous agent.*

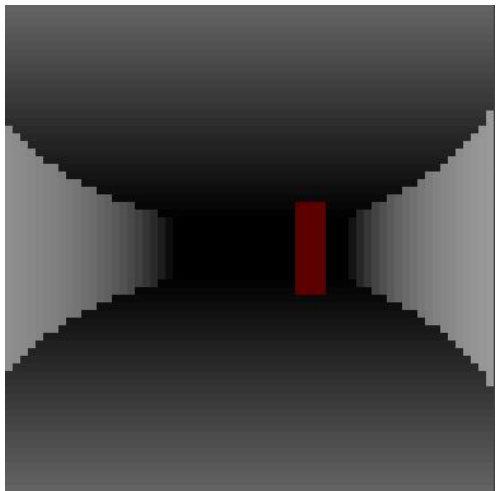

*Figure 3: The transformed 3D environment resulting from the application of ray tracing techniques to the initial 2D space.*

Figure 2 depicts the 2D space representative of the initial environment where the agent operated. This space acts as the canvas for capturing data from the simulated LiDAR sensor and camera for subsequent processing. Figure 3 illustrates the transformed 3D image resulting from the application of ray tracing techniques to the 2D space.

### C. Training Process

The agent's model is trained through the process of imitation learning. To achieve this process, there are five controls that can be switched on/off or activated during the training process. These controls are:

1. Recording – The sensor data and the trainer's current action at the current state are added to the dataset.
2. Training – The model's weights are fit for one epoch to the current data. The data is shuffled each time this is called.
3. Autonomous Mode – The agent continuously takes in the environment data and performs the top predicted action. This is used to evaluate agent performance when training.
4. Save Model – The current weights and model architecture are saved to the directory for later use.
5. Load Model – An old model selected from the directory will be loaded into the agent for further training, tuning, and testing.

Utilizing these controls allows training to be performed as follows:

1. Record behaviors into the dataset.
2. Tune model on current data.
3. Run agent in autonomous mode in various parts of the environment to evaluate performance.
4. Record additional data and/or tune model on dataset.
5. Save model for evaluation of generalization.

### D. Evaluation Metrics

Our model was evaluated using a monte-carlo simulation, which tests the agent by initializing it in randomized environments. The agent would then autonomously navigate until it successfully reaches the goal, collides with a wall, or runs out of time. Many episodes are used to get an average success rate of the model. Each episode is defined by a uniquely generated map with a unique goal location. Successful instances of target identification were recorded, while cases of the robot passing by the target within sensor range without recognizing it and continuing its search were considered failures. The time taken for the robot to seek and find the target in each episode was documented, and the average time was computed by taking the mean across all episodes. The accuracy of the model was defined and calculated as the ratio of successes to the total number of runs, as represented by (1).

$$Accuracy = \frac{Success}{no. of\ runs} \quad (1)$$

## V. EXPERIMENTAL DESIGN

### A. Computation Resources

The training and testing of deep learning models and other computations carried out in this project were made possible through the utilization of the computational resources listed in Table 1.

*Table 1: Computational resources specifications of the machines used in this project*

| Processor | 13$^{th}$ Gen Intel i9-13900K |
|---|---|
| RAM | 32 GB DDR5 |
| GPU | NVIDIA Geforce RTX 3080 |
| VRAM | 10GB GDDR6X |

### B. Hyperparameter Tuning

## VI. RESULTS

*Table 2: Accuracy report of the model tested on three maps and an average of the corresponding time steps to end simulations*

| Map | Accuracy | Average Time |
|---|---|---|
| Test 1 | 85% | 1726 steps |
| Test 2 | 85% | 1103 steps |
| Test 3 | 80% | 1189 steps |

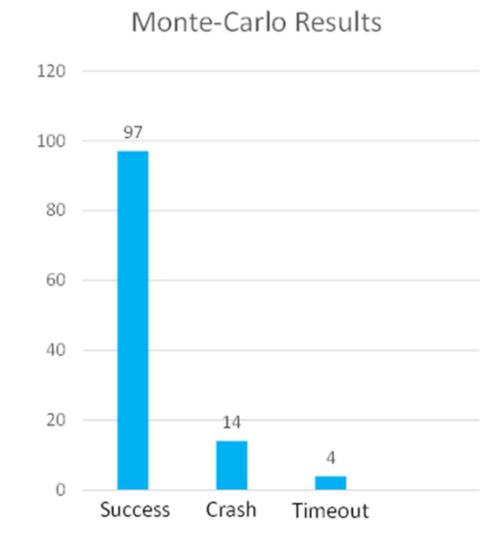

*Figure 4: Example Evaluation Results.*

## VII. DISCUSSION AND CONCLUSION

This proposed network fusion model can autonomously navigate complex and unknown environments with high accuracy. A simulation of a subterranean environment is constructed along with sensor models to show the efficacy of our trained model in practice.

The authors show that imitation learning is capable of training complex behaviors within a larger multimodal network structure. Training showed and flexibility were maintained with behavior tuning.

Overall, the model generalized well from imitation learning; it scored above 80% accuracy for the randomized set of trials. Some common causes for model failure, as defined above, are as follows:

• The goal spawns behind the agent and the agent runs out of time progressing through a long course and returning to the start.
• The goal spawns in the corner and the agent will turn out of the corner before reaching the goal.
• The goal is in sight of the agent and the agent mistakenly crashes into the wall on the way to the goal or right next to the goal.

This behavior can be attributed to the last model trained and evaluated. The initial course ensured that the goal was positioned in the center of the hallway. This means that there was not any training data relating to these unique circumstances of a hard-to-reach goal location, such as in a corner or against a wall.

The success of the work depends greatly on the user effectively training the model for desired behavior. This entails providing a rich dataset of actions with correct corresponding behavior for all actions. Minimizing recordings of contradicting behaviors in the learning dataset directly correlates to robust decision making. The model was found to get stuck in an alternating pattern of behavior due to inconsistencies learned in earlier iterations.

Finally, the performance of the model on the recorded training dataset is not an important aspect of the model's actual performance because the Monte Carlo evaluation is the only way to evaluate whether the agent has learned all desired behaviors. Approximately 15 iterations of manual training were used to refine the tuning/recording process and evaluation metrics.

## VIII. FUTURE WORK

In our ongoing deep learning project, the core objective involves training within a simulation environment, an autonomous navigation agent tasked with the challenge of autonomously locating a designated target. However, as we transitioned into the testing phase, we encountered notable limitations. The seeking behavior of the trained model was observed to be relatively slow when navigating through unfamiliar environments, and its overall performance demonstrated shortcomings in adapting to more intricate and complex settings. In light of these observations, our attention turns to future work where we aim to comprehensively address these issues.

One key area of focus involves improving the speed of the model. We plan to implement optimization strategies and explore algorithmic enhancements to expedite decision-making and action execution. Concurrently, we will prioritize the augmentation of the model's generalization capabilities in complex environments by enriching the training dataset with diverse scenarios and delving into advanced architectural modifications. Through these strategic initiatives, our goal is to cultivate a more agile, adaptable, and high-performing autonomous navigation system capable of excelling across a spectrum of challenging real-world scenarios.

Additionally, we propose improving the imitation learning generalization by augmenting the dataset with training iterations that vary the locations of the goalpost. Goalpost location variances in training allow the trainer to record troublesome situations, such as reaching a goalpost located in a corner, while also improving simple, already learned behavior regarding navigation and reaching a goalpost located in central locations. Generally, this should provide a richer dataset for the model to learn from and, as a result, improve the model's decision ability.

Additional extensions to the project may allow the model to potentially increase performance. To elaborate, we propose adding a memory component such as a map, generated by SLAM. The raw data would be processed via a convolutional network similar to the image processing channel before being incorporated into the controller model architecture.

Additionally, we can implement reinforcement learning (RL) in the future to further improve the model's ability to generalize in a decision-making environment. Because the model is dependent on the user training the model effectively, undesired behavior can be taught against while simultaneously training desired behavior. This can be done by implementing RL where the agent is punished for crashing and rewarded for reaching the goal (and potentially additionally rewarded for reaching the goal more quickly). Such a system would benefit from the desired pretrained weights compiled from imitation learning, due to the complexity of the model, which implements multimodal data and must learn multiple behaviors at the same time.